\documentclass[wcp]{jmlr}

\usepackage{longtable}

\usepackage{booktabs}

\usepackage{lineno}

\usepackage{graphicx}
\usepackage{comment}
\usepackage{amsmath,amssymb}
\usepackage{multirow}
\usepackage{booktabs}
\usepackage{makecell}
\usepackage{hhline}
\usepackage{stfloats} 
\usepackage{algorithm}
\usepackage{algorithmic}
\usepackage{multicol}
\usepackage{subcaption}
\usepackage{wrapfig}

\makeatletter
\let\Ginclude@graphics\@org@Ginclude@graphics 
\makeatother

\jmlrvolume{222}
\jmlryear{2023}
\jmlrworkshop{ACML 2023}

\title[Patch-level Neighborhood Interpolation~(Pani)]{Patch-level Neighborhood Interpolation: A General and Effective Graph-based Regularization Strategy}

  \author{\Name{Ke Sun$^*$} \Email{ajksunke@pku.edu.cn}\\
  \addr Center for Data Science, Peking University
  \AND
	\Name{Bing Yu$^*$} \Email{byu@pku.edu.cn}\\
	  \addr School of Mathematical Sciences, Peking University
  \AND
	\Name{Zhouchen Lin} \Email{zlin@pku.edu.cn}\\
\addr 1. National Key Lab of General AI, School of Intelligence Science and Technology, Peking University, 
\addr 2. Institute for Artificial Intelligence, Peking University, 
\addr 3. Peng Cheng Laboratory
\AND
\Name{Zhanxing Zhu$^{\dagger}$} \Email{zhanxing.zhu@pku.edu.cn}\\
 \addr School of Mathematical Sciences, Peking University}

\editors{Berrin Yan{\i}ko\u{g}lu and Wray Buntine}

\begin{document}

\maketitle

\def\thefootnote{*}\footnotetext{The two authors contributed equally to this work}\def\thefootnote{\arabic{footnote}}

\def\thefootnote{$\dagger$}\footnotetext{Corresponding author}\def\thefootnote{\arabic{footnote}}

\begin{abstract}
Regularization plays a crucial role in machine learning models, especially for deep neural networks. The existing regularization techniques mainly rely on the i.i.d. assumption and only consider the knowledge from the current sample, without the leverage of the neighboring relationship between samples. In this work, we propose a general regularizer called \textbf{Patch-level Neighborhood Interpolation~(Pani)} that conducts a non-local representation in the computation of networks. Our proposal explicitly constructs patch-level graphs in different layers and then linearly interpolates neighborhood patch features, serving as a general and effective regularization strategy. Further, we customize our approach into two kinds of popular regularization methods, namely Virtual Adversarial Training (VAT) and MixUp as well as its variants. The first derived \textbf{Pani VAT} presents a novel way to construct non-local adversarial smoothness by employing patch-level interpolated perturbations. The second derived \textbf{Pani MixUp} method extends the  MixUp, and achieves superiority over MixUp and competitive performance over state-of-the-art variants of MixUp method with a significant advantage in computational efficiency. Extensive experiments have verified the effectiveness of our Pani approach in both supervised and semi-supervised settings.
\end{abstract}

\section{Introduction}

In the statistical learning theory, regularization techniques are typically leveraged to achieve the trade-off between empirical error minimization and the control of model complexity~\citep{vapnik2015uniform}. In contrast to the classical convex empirical risk minimization where regularization can rule out trivial solutions, regularization plays a rather different role in deep learning due to its highly non-convex optimization nature~\citep{zhang2016understanding}. Among all the explicit and implicit regularization, regularization with stochastic transformation, perturbations and randomness, such as adversarial training~\citep{goodfellow2014explaining}, dropout and MixUp~\citep{zhang2017mixup}, play a key role in the deep learning models due to their superiority in the performance~\citep{berthelot2019mixmatch,zhang2017mixup,miyato2018virtual,berthelot2019remixmatch, yao2022c, kim2021co}. 

Nevertheless, the vast majority of regularization methods, including the aforementioned ones, assume that the training samples are drawn independently and identically from an unknown data-generating distribution. However, this i.i.d. assumption is commonly violated in realistic scenarios where batches or sub-groups of training samples are likely to have internal correlations. Accounting for the correlations in real-world training data can lead to statistically significant improvements in accuracy~\citep{dundar2007learning, yun2019cutmix} and other benefits, including the robustness~\citep{svoboda2018peernets, yao2022c}. For example, Peer-Regularized Networks~(PeerNet)~\citep{svoboda2018peernets} applied graph convolutions~\citep{velivckovic2017graph,kipf2016semi} to harness information of peer samples, and verified its effectiveness on defending adversarial attacks. Motivated by these facts, we design a general regularization strategy called Patch-level Neighborhood Interpolation~(Pani) that can fully utilize the internal relationship between samples by explicitly constructing a graph within a mini-batch in order to consistently improve the generalization of deep neural networks in both semi- and supervised settings. Our contributions can be summarized as folllows:

\begin{itemize}
	\item We propose a simple yet general effective non-local regularization called \textbf{Patch-level Neighborhood Interpolation~(Pani)}. Pani linearly interpolates on the neighboring patch features and yields a non-local representation that additionally captures the relationship of neighboring patch features in different layers.
	
	\item We explicitly customize our Pani approach into two classes of popular and general regularization strategies, i.e., Virtual Adversarial Regularization and MixUp, resulting in \textbf{Pani VAT} and \textbf{Pani MixUp}. Pani VAT yields non-local adversarial perturbations, providing a more informative adversarial smoothness in the semi-supervised learning setting. Pani MixUp and MixMatch perform better than their vanilla versions with computational efficiency by mixing fine-grained patch features and supervised signals.
	
	\item Extensive experiments demonstrate that both of the two derived regularization strategies can outperform other state-of-the-art approaches in both supervised and semi-supervised tasks, presenting the generality and superiority of our Pani method.
\end{itemize}

\subsection{Related Work}\label{sec:relatedwork}


\paragraph{Virtual Adversarial Training~(VAT) and Its Variants.} Adversarial Training~\citep{goodfellow2014explaining,madry2017towards,zhang2019theoretically, tsipras2018robustness} can provide a new form of regularization beyond that provided by other generic regularization strategies, such as dropout, pretraining and model averaging. VAT~\citep{miyato2018virtual} extends the adversarial training through adversarially smoothing the posterior output distribution with the leverage of unlabeled data,  achieved great success in image classification~\citep{miyato2018virtual}, text classification~\citep{miyato2016adversarial} and node classification~\citep{sun2019virtual}. There is a flurry of VAT variants~\citep{luo2017smooth,yu2019tangent}, most of which heavily rely on generative models to construct data manifold. Tangent-Normal Adversarial Regularization~(TNAR)~\citep{yu2019tangent} extended VAT by considering the data manifold and applied VAT along the tangent space and the orthogonal normal space of the data manifold, outperforming previous semi-supervised approaches. VAT+SNTG~\citep{luo2017smooth} constructed a graph based on the predictions of the teacher model to smooth the representation on the low-dimensional manifold in the semi-supervised setting. By contrast, our Pani method is a fine-grained patch-level and more general regularization that can be leveraged to refine the representation of deep neural networks in both semi- and supervised scenarios \textit{without the requirement of any generative model.}

\paragraph{MixUp and Its Variants.} MixUp~\citep{zhang2017mixup} has been widely used in various machine learning tasks. However, MixUp samples tend to be unnatural and locally ambiguous, which may confuse the model~\citep{yun2019cutmix}. To overcome this issue, CutMix~\citep{yun2019cutmix} replaces the image region with a patch from another training sample in a more straightforward way. MixUp variants, including Manifold MixUp~\citep{verma2019manifold}, AdaMixUp~\citep{guo2019mixup}, SmoothMix~\citep{lee2020smoothmix},  Puzzle Mix~\citep{kim2020puzzle} and  Co-MixUp~\citep{kim2021co},   have been proposed through various kinds of interpolation and transformations. In addition, C-MixUp~\citep{yao2022c} extends MixUp into the regression setting, investigating both its improvement on in-distribution generalization and out-of-distribution robustness. In semi-supervised learning, FixMatch~\citep{sohn2020fixmatch} and MixMatch~\citep{berthelot2019mixmatch}  serve as a natural extension of MixUp and achieve state-of-the-art accuracy. In contrast with VAT, MixMatch~\citep{berthelot2019mixmatch} utilizes one specific form of consistency regularization, i.e., using the standard data augmentation for images, such as random horizontal flips, rather than computing adversarial perturbations to smooth the posterior distribution of the classifier. Our Pani method is proposed differently from other MixUp variants, and shares similarities with CutMix. However, CutMix randomly cut patches, while our Pani linearly interpolates patch features and is more general. Pani Method can also easily enhance the performance of MixMatch in the semi-supervised setting.

\paragraph{Manifold Regularization.} There is a flurry of papers introducing regularization from classical manifold learning based on the assumption that the data can be modeled as a low-dimensional manifold in the data space. As demonstrated in\citep{hinton2012improving, ioffe2015batch}, regularizers that work well in the input space can also be applied to the hidden layers of a deep network, which could further improve the generalization performance. Our Patch-level Neighborhood Interpolation can be easily extended from input to the hidden layers, enjoying the benefits of manifold regularization.

\paragraph{Non-local Image Filtering.} Past non-local image filter methods~\citep{tomasi1998bilateral,buades2005non,sochen1998general} leveraged both the pixel intensities and their pixel neighbors together with their locations to design these non-shift-invariant filters. Recently, Non-local Neural Networks~\citep{wang2018non} presented one effective non-local operation that serves as a generic component for capturing long-range dependencies with deep neural networks. Similarly, our Pani still can capture the correlation knowledge of patch features within a batch, therefore yielding an improvement in performance for the derived methods. Moreover, our method also serves as a novel non-i.i.d. regularization and can reasonably generalize well to broader settings especially when the natural correlation in the sub-group exists.

\section{Preliminary}\label{sec: preliminary}

\paragraph{Virtual Adversarial Training~(VAT).} VAT replaces true labels $y$ of samples in the formulation of adversarial training by current estimate $p(y|x;\hat{\theta})$ from the model: 
\begin{equation} \begin{aligned} 
		\min_{\theta} \max_{r,\|r\|\leq\epsilon} D\left[p(y|x;\hat{\theta}), p(y|x+r;\theta)\right],
	\end{aligned} \label{eq_vat1} \end{equation}
where $D[q,p]$ measures the divergence between two distributions $q$ and $p$. $r$ is the adversarial perturbation depending on the current sample $x$ that can further provide smoothness in SSL. Then the VAT regularization $\mathcal{R}_{\rm vadv}(x,r;\theta)$ is derived from the inner maximization:
\begin{equation}\begin{aligned} 
		\mathcal{R}_{\rm vadv}(x,r;\theta) =  \max_{r,\|r\|\leq\epsilon} D\left[p(y|x;\hat{\theta}), p(y|x+r;\theta)\right]
	\end{aligned} \label{eq_vat2} \end{equation}
One elegant part of VAT is that it utilized the second-order Taylor's expansion of virtual adversarial loss to compute the perturbation $r$, which can be computed efficiently by power iteration with finite differences. Once the desired perturbation $r^*$ has been obtained, we can conduct forward and back propagation to optimize the full loss function:

\begin{equation}
	\min_{\theta} \mathcal{L}_0 +  \beta \mathbb{E}_{x\sim \mathcal{D}}\mathcal{R}_{\rm vadv}(x,r^*;\theta), 
	\label{eq_vat3} \end{equation}
where $\mathcal{L}_0$ is the original supervised loss and $\beta$ is the hyper-parameter to control the degree of virtual adversarial smoothness. 

\paragraph{MixUp.} MixUp~\citep{zhang2017mixup} augments the training data with linear interpolation on both input features and target. The resulting feature-target vectors are shown as follows:
\begin{equation}
	\begin{aligned}
		\tilde{x} = \lambda x_i + (1 - \lambda) x_j, \quad  \tilde{y} = \lambda y_i + (1 - \lambda) y_j,
	\end{aligned}
\end{equation}
where $(x_i, y_i)$ and $(x_j, y_j)$ are two feature-target vectors drawn randomly from the training data. $\lambda \sim \text{Beta}(a, a)$ and $a \in (0, \infty)$. MixUp can be understood as a form of data augmentation that encourages decision boundaries to transit linearly between classes. It is a kind of generic regularization that provides a smoother estimate of uncertainty, yielding the improvement of generalization. 

\paragraph{Peer-Regularized Networks~(PeerNet).} The centerpiece of PeerNet~\citep{svoboda2018peernets} is the learnable \textit{Peer Regularization}~(PR) layer designed to focus on improving the adversarial robustness of deep neural networks. PR layer can be flexibly added to the feature maps of deep models.  Let $\mathbf{Z}^1,...,\mathbf{Z}^N$ be $n \times d$ matrices as the feature maps of $N$ images, where $n$ is the number of pixels and $d$ represents the dimension of the feature in each pixel, i.e., number of channel in the feature map. The core of PeerNet is to find the $K$ nearest neighboring pixels for each pixel among all the pixels of $N$ peer images via constructing a $K$ nearest neighbor graph in the $d$-dimensional  space. Particularly, for the $p$-th pixel in the $i$-th image $\mathbf{z}^i_p$, the $k$-th nearest pixel neighbor can be denoted as $\mathbf{z}^{j_k}_{q_k}$ taken from the  pixel $q_k$ of the  peer image $j_k$. Then the learnable PR layer is constructed by a variant of Graph Attention Networks~(GAT)~\citep{velivckovic2017graph}:
\begin{equation}
	\begin{aligned}
		\tilde{\mathbf{z}}_{p}^{i}=&\sum_{k=1}^{K} \alpha_{i j_{k} p q_{k}} \mathbf{z}_{q_{k}}^{j_{k}},  \quad \alpha_{i j_{k} p q_{k}}=\frac{\text { LeakyReLU }\left(\exp \left(f_a\left(\mathbf{z}_{p}^{i}, \mathbf{z}_{q_{k}}^{j_{k}}\right)\right)\right)}{\sum_{k^{\prime}=1}^{K} \text { LeakyReLU }\left(\exp \left(f_a\left(\mathbf{z}_{p}^{i}, \mathbf{z}_{q_{k^{\prime}}}^{j_{k^{\prime}}}\right)\right)\right)},
	\end{aligned}
\end{equation}
where $\alpha_{i j_{k} p q_{k}}$ is the attention score determining the importance of the $q_k$-th pixel of the $j$-th peer image on the representation of current $p$-th pixel $\tilde{\mathbf{z}}^i_p$ taken from the $i$-th image. $f_a()$ is a fully connected layer mapping from 2$d$-dimensional input to scalar output. Therefore, the resulting learnable PR layer involves non-local filtering by leveraging the wisdom of pixel neighbors from peer images, showing robustness against adversarial attacks.

\section{Our Method: Patch-level Neighborhood Interpolation}

For our method, one related work is PeerNet~\citep{svoboda2018peernets} that designed graph-based layers to defend against adversarial attacks, but unfortunately, the construction of pixel-level $K$-NN graphs in PeerNet~\citep{svoboda2018peernets} is costly in computation. By contrast, our motivation is to develop a general regularization that can consistently boost the performance of deep neural networks in both semi- and supervised settings rather than the adversarial scenario. Besides, the construction way of a non-local layer in our method is more flexible and can be determined by the specific objective function, as elaborated in the following part. Moreover, our patch-level method can achieve the computational advantage over pixel-level regularization, and incorporates more meaningful semantic correlations in different layers. Particularly, a flexible patch size can be chosen according to the size of the receptive field in different layers, yielding a more informative graph-based representation and better regularization performance.

\begin{figure*}[b!]
	\centering	\centering\includegraphics[width=0.7\textwidth,trim=100 130 60 85,clip]{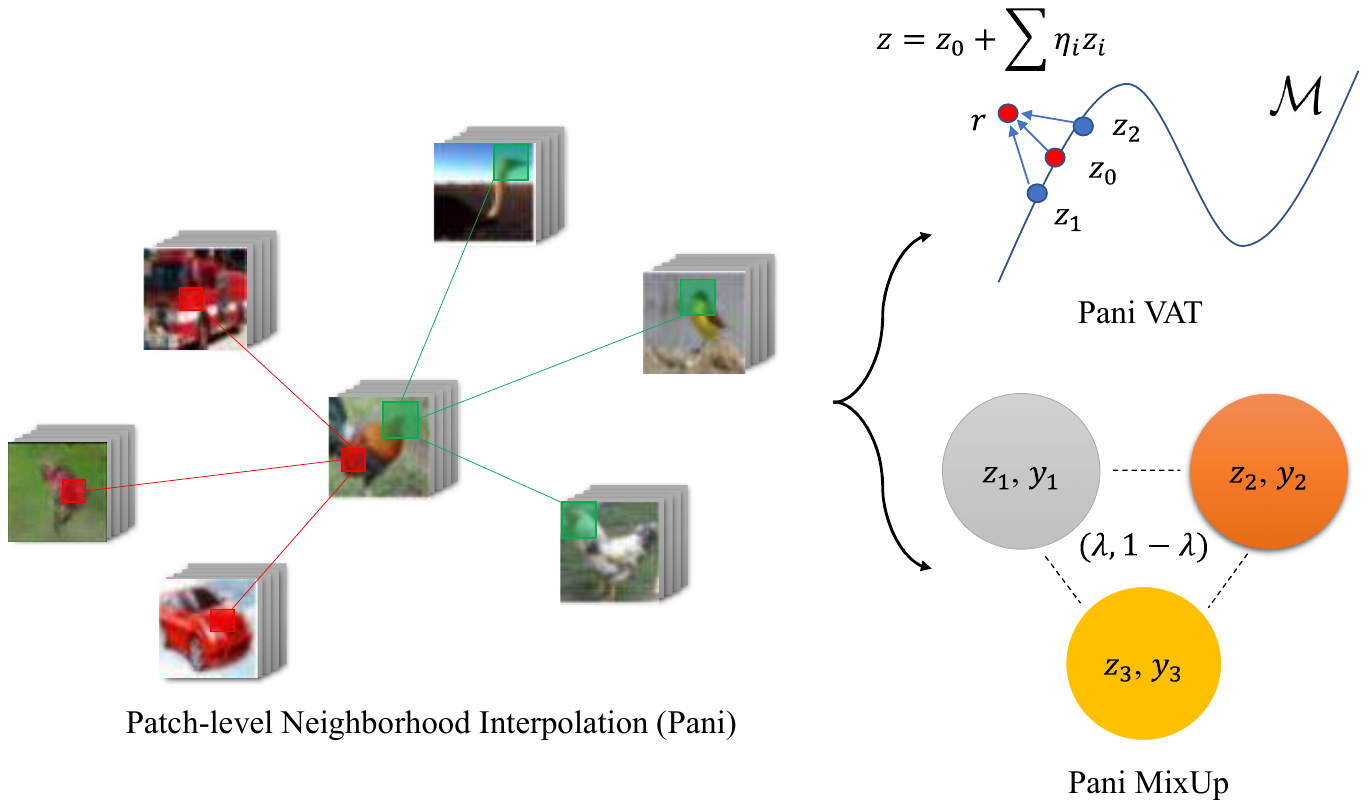}
	\caption{Pipeline of our Patch-level Neighborhood Interpolation followed by two derived regularizations, i.e., Pani VAT and Pani MixUp. $r$ represents the perturbation constructed by our method and $(\lambda, 1-\lambda)$ is the mixing coefficient pair.}
	\label{figure_framework}
\end{figure*}

\paragraph{Step 1: Construction of $K$ nearest neighbor patch graphs.} As illustrated in Figure~\ref{figure_framework}, in the first step of Patch-level Neighborhood Interpolation (Pani) we determine the candidate peer images set $\mathcal{S}_i$ for each image $i$. This can be achieved by random matching or computing the semantically nearest image neighbors using e.g. the cosine distance. Next, we construct the whole patches set $\mathcal{P}_i$ on the candidate peer images set $\mathcal{S}_i$ for each image $i$ by clipping the corresponding patches in the different locations on an input or a feature map. Following the establishment of patch set  $\mathcal{P}_i$, we construct $K$ nearest neighbor patch graphs based on the distance of patch features in order to find the neighbors of each patch in a patch set $\mathcal{P}_i$ for $\forall i=1,..,N$. Mathematically, following the definition in the PeerNet, let $\mathbf{z}^i_p$ be the $p$-th patch on the input or feature map $\mathbf{Z}^i$ for the $i$-th image within one batch. Then denote the $k$-th nearest patch neighbor for $\mathbf{z}^i_p$ as $\mathbf{z}^{j_k}_{q_k}$ taken from the  patch $q_k$ of the peer image  $j_k$ in the candidate set $\mathcal{S}_i$. 

\paragraph{Step 2: Linear interpolation.} Next, in order to leverage the knowledge from neighbors, different from the graph attention mechanism in PeerNet, we apply a more straightforward linear interpolation on the neighboring patches for the current patch $\mathbf{z}^i_p$. Then, the general formulation of our Patch-level Neighborhood Interpolation can be presented as follows:
\begin{equation}
	\begin{aligned}
		\tilde{\mathbf{z}}^i_p = \mathbf{z}^i_p + \sum_{k=1}^{K}\eta_{ipk} (\mathbf{z}^{j_k}_{q_k} - \mathbf{z}^i_p),
	\end{aligned}
	\label{eq_framework}
\end{equation}
where $\eta_{ipk}$ is the combination coefficient for the $p$-th patch of $i$-th image w.r.t its $k$-th patch neighbor, which can be computed through the power iteration similar to the manner of VAT, or through random sampling from a specific distribution in randomness-based regularization, e.g., MixUp and its variants. Moreover, the choice of linear interpolation in Eq.~\ref{eq_framework} enjoys a computational advantage over the nonlinear GAT form in PeerNet in the computation of networks. Finally, after the patch-level linear interpolation on patch features, we can obtain the refined graph-based representation $\tilde{\mathbf{Z}}^i$ for $i$-th image, $\forall i=1,...,N$.  Note that our proposed method can explicitly combine the advantages of manifold regularization and non-local filtering in a flexible way. To further demonstrate the generality and effectiveness of our Pani method, we propose the Pani version of two typical regularization strategies, i.e., VAT and MixUp as well its variant MixMatch.

\subsection{Pani VAT}\label{section:Pani_VAT}
Based on our Pani framework, we can construct a novel Pani VAT that utilizes the linear interpolation of patch neighbors for each sample to manipulate the non-local perturbations, thus providing more informative adversarial smoothness in the semi-supervised setting. Consider a more general composite function form of the classifier $f$, i.e.,$f(x)=g(z)$ and $z=h(x)$ where $z$ denotes the hidden feature of input $x$ or the input itself when the reduced form happens. Combining VAT formulation Eq.~\ref{eq_vat2}, and Pani formulation Eq.~\ref{eq_framework}, we reformulate Pani VAT with perturbations on $L$ layers in a deep neural network as follows:
\begin{equation}
	\begin{aligned}
		& \ \max_{\eta} D[g(z), g(\tilde{z}(\eta))] \ \ s.t.  \ \ \sum_{l=1}^{L} w^2_l  \Vert \eta^{(l)} \Vert ^2 \le \epsilon^2,\\
	\end{aligned}
	\label{eq_NVAT}
\end{equation}
where $D$ measures the divergence between two distributions. $\eta=\{\eta_{ipk}\}$ denotes the generic perturbations from our Pani method and $\eta^{(l)}$ indicates the perturbations in $l$-th layer of network. $\tilde{z}(\eta)=\{\tilde{\mathbf{z}}^i_p\}$ represents the smoothed feature map imposed by perturbation $\eta$ considering all patches in the way shown in Eq.~\ref{eq_framework}. In particular, when $L=1$, adversarial perturbations are only imposed on the input feature, which is similar to the traditional virtual adversarial perturbations. Additionally, $w_l$ is the hyper-parameter, adjusting the weight of perturbation $\eta^{(l)}$ in different layers with the overall perturbations restrained in an $\epsilon$-ball. Next, we still utilize the similar power iteration and finite difference proposed in VAT to compute the desired perturbation $\eta^*$. The resulting full loss function is defined as:
\begin{equation}
	\min_{\theta} \mathcal{L}_0 +  \beta\mathbb{E}_{x\sim \mathcal{D}}\mathcal{R}_{\rm vadv}(x,\eta^*;\theta), 
	\label{eq_NVAT2}
\end{equation}
where $\mathcal{L}_0$ is the original supervised loss and $\beta$ controls the degree of adversarial smoothness. $\mathcal{R}_{\rm vadv}(x,y,\eta^*)= D[g(z), g(\tilde{z}(\eta^*))]$ can be attained after solving the optimization problem in Eq.~\ref{eq_NVAT}. For the implementation details, we describe them in Algorithm~\ref{alg:PaniVAT}.

\begin{algorithm}[t!]
	\caption{Pani VAT within a Batch}\label{alg:lanczos}
	\begin{algorithmic}[1]
		\STATE \textbf{Input:} neighbors $K_1$, $K_2$, classifier $f$, batch size $B$, perturbed layers $L$
		\STATE \textbf{Initialization:} combination coefficient $\eta$
		\STATE Compute $K_1$ nearest image neighbors based on the distance of the second last layer output from $f$ and obtain $K_1$~($K_1 \le B$) peer images set $\mathcal{S}_i$ for each image $i$.
		\STATE \textbf{for} $l=1$ \textbf{to} $L$ \textbf{do}: 
		\STATE \quad Compute the patch set $\mathcal{P}_i$ for all $K_1$ peer images on layer $l$  for each image $i$ .
		\STATE \quad Construct a $K_2$ nearest patch neighbors graph for each patch in each image $i$.
		\STATE \quad Conduct Patch-level Neighborhood Interpolation via Eq.~\ref{eq_framework} for each patch.
		\STATE \textbf{end for}
		\STATE Conduct power iteration and finite difference to compute $\eta^*$ constrained by Eq.~\ref{eq_NVAT}.
		\STATE \textbf{Return} $\mathcal{R}_{\rm vadv}(x,\eta^*;\theta)$
	\end{algorithmic}
	\label{alg:PaniVAT}
\end{algorithm}

\paragraph{Remark.} As shown in the adversarial part of Figure~\ref{figure_framework}, the rationality of our Pani VAT method lies in the fact that the constructed perturbations can entail more non-local information coming from neighbors of the current sample. Through the delicate patch-level interpolation among neighbors of each patch, the resulting non-local virtual adversarial perturbations are expected to provide more informative smoothness, thus enhancing the performance of the classifier in the semi-supervised setting.

\subsection{Pani MixUp}\label{section:Pani_Mixup}

Next, we leverage Patch-level Neighborhood Interpolation to derive Pani MixUp. The core formulation of Pani MixUp can be written as:
\begin{equation}
	\begin{aligned}
		&\tilde{\mathbf{z}}^i_p = (1-\sum_{k=1}^{K}\eta_{ipk})\mathbf{z}^i_p + \sum_{k=1}^{K}\eta_{ipk} \mathbf{z}^{j_k}_{q_k}\\
		&\tilde{y}_i = (1- \sum_{k=1}^{K}\sum_{p=1}^{P}\frac{\eta_{ipk}}{P}) y_i + \sum_{k=1}^{K}\sum_{p=1}^{P}\frac{\eta_{ipk}}{P} \  y_{j_k}, \ \ s.t. \  \lambda = 1- \sum_{k=1}^{K}\sum_{p=1}^{P}\frac{\eta_{ipk}}{P} , \\
	\end{aligned}
	\label{eq_NMU}
\end{equation}
where $(\mathbf{z}^i,y^i)$ are the feature-target pairs randomly drawn from the training data. $P$ is the number of patches in each image and $\lambda \sim \text{Beta}(a,b)$ represents the importance of the current input or target while conducting MixUp. To compute $\eta_{ipk}$, we firstly sample $\lambda$ from $\text{Beta}(a,b)$ and $\eta_{ipk}^0$ from a uniform distribution respectively, then we normalize $\eta_{ipk}^0$ according to the ratio of $\lambda$ to satisfy the constraint in Eq.~\ref{eq_NMU} and thus obtain $\eta_{ipk}$. It should be noted that due to the unsymmetric property of $\lambda$ in our framework, we should tune both $a$ and $b$ in our experiments. For simplicity, we fix $b=1$ and only consider the $a$ as the hyper-parameter to pay more attention to the importance of the current patch, which is inspired by the similar approach in MixMatch~\citep{berthelot2019mixmatch}. Here we reformulate Eq.~\ref{eq_NMU} to illustrate that Pani MixUp is naturally derived from our Pani framework by additionally considering several constraints:
\begin{equation}
	\begin{aligned}
		&\tilde{\mathbf{z}}^i_p = \mathbf{z}^i_p + \sum_{k=1}^{K}\eta_{ipk} (\mathbf{z}^{j_k}_{q_k} - \mathbf{z}^i_p)\\
		&s.t. \  \lambda = 1 - \sum_{k=1}^{K}\sum_{p=1}^{P}\frac{\eta_{ipk}}{P}, \forall i = 1,...,N,  \lambda \sim \text{Beta}(a,b), \eta_{ipk} \in [0,1], \forall i,p,k
	\end{aligned}
	\label{eq_NMU2}
\end{equation}
where the first constraint in Eq.~\ref{eq_NMU2} can be achieved through normalization via $\lambda$. Meanwhile, we impose $\eta_{ipk} \in [0,1]$ as $\eta_{ipk}$ represents the interpolation coefficient. Further, we elaborate on the procedure of Pani MixUp in Algorithm~\ref{alg:PaniMixUp}.

\begin{algorithm}[htbp]
	\caption{Pani MixUp within a Batch}
	\begin{algorithmic}[1]
		\STATE \textbf{Input:} neighbors $K$, classifier $f$, batch size $B$, perturbed layers $L$, parameter $a$
		\STATE Compute peer images by random matching and obtain peer images set $\mathcal{S}_i$ for each image $i$.
		\STATE \textbf{for} $l=1$ \textbf{to} $L$ \textbf{do}: 
		\STATE \quad Compute the patch set $\mathcal{P}_i$ on layer $l$  for each image $i$.
		\STATE \quad Construct a $K$ nearest patch neighbors graph for each patch in each image $i$.
		\STATE \quad Sample initial coefficient $\eta^{(l)}_0=\{\eta_{ipk}^0\}$ from $U(0, 1)$ and $\lambda$ from $\text{Beta}(a,1)$.
		\STATE \quad Normalize $\eta^{(l)}_0$ according to the ratio $\lambda$ via Eq.~\ref{eq_NMU2} to compute $\eta^{(l)}$.
		\STATE \quad Conduct Pani MixUp over patch features and labels via Eq.~\ref{eq_NMU2} for each patch.
		\STATE \textbf{end for}
		\STATE \textbf{Return} supervised loss based on mixed features and labels.
	\end{algorithmic}
	\label{alg:PaniMixUp}
\end{algorithm}

\paragraph{Remark.} Different from the role of $\eta$ in the aforementioned Pani VAT where $\eta$ serves as the interpolated perturbations, the physical meaning of $\eta$ in our Pani MixUp approach is the linear interpolation coefficient to conduct MixUp. Despite this distinction, both of the two extended regularization methods are naturally derived from our Pani framework, further demonstrating the generality and superiority of our Pani strategy.

\section{Experiments}
In this section, we conduct extensive experiments for Pani VAT and Pani MixUp and its variant Pani MixMatch on both semi- and supervised settings.

\subsection{Pani VAT}

\begin{table}[t!]
	\centering
	\scalebox{0.65}{
		\begin{tabular}{lcc}
			\hhline{===}
			Method  & CIFAR-10~(4,000 labels) \\
			\hline
			VAT~\citep{miyato2017virtual} & $13.15\pm 0.2$ \\
			VAT + SNTG~\citep{luo2017smooth} & $12.49\pm 0.36$ \\
			$\Pi$ model~\citep{laine2016temporal} & $16.55\pm 0.29$ \\
			Mean Teacher~\citep{tarvainen2017mean}  & $17.74\pm 0.30$ \\
			CCLP~\citep{kamnitsas2018semi}  & $18.57\pm 0.41$ \\
			ALI~\citep{dumoulin2016adversarially}& $17.99\pm 1.62$ \\
			Improved GAN~\citep{salimans2016improved} & $18.63\pm 2.32$ \\
			Tripple GAN~\citep{li2017triple}  & $16.99\pm 0.36$ \\
			Bad GAN~\citep{dai2017good}  & $14.41\pm 0.30$ \\
			LGAN~\citep{qi2018global}  & $14.23\pm 0.27$ \\
			Improved GAN + JacobRegu + tangent~\citep{kumar2017semi} &$16.20\pm 1.60$ \\
			Improved GAN + ManiReg~\citep{lecouat2018semi} & $14.45\pm 0.21$ \\
			TNAR (with generative models)~\citep{yu2019tangent}  & $12.06 \pm 0.35$ \\
			\hline
			Pani VAT (input)& $12.33 \pm 0.091$ \\
			Pani VAT (+hidden)& $\mathbf{11.98 \pm 0.106}$ \\
			\hhline{===}
		\end{tabular}
	}
	\vspace{1.0ex}
	\caption{Classification errors ($\%$) of compared methods on CIFAR-10 dataset without data augmentation. The results of our Pani methods are the average ones under 4 runs.}
	\label{table_vat}
\end{table}

\paragraph{Implementation Details.} For fair comparison with VAT and its variants, e.g., VAT + SNTG~\citep{luo2017smooth} and TNAR~\citep{yu2019tangent}, we choose the standard large convolutional network as the classifier as in \citep{miyato2018virtual}. For the option of dataset, we focus on the standard semi-supervised setting on CIFAR-10 with 4,000 labeled data. Unless otherwise noted, all the experimental settings in our method are the identical with those in the Vanilla VAT~\citep{miyato2018virtual}. In particular, we conduct our Pani VAT on input layer and one additional hidden layer, yielding  two variants Pani VAT~(input) and Pani VAT~(+hidden). \textit{For the option of hyper-parameters, we conduct the delicate line search for the best performance.} In Pani VAT~(input), we choose patch size as 2, $K_1=10$ for the number of peer images, $K_2=10$ to construct the nearest patch neighbor graph, perturbation size $\epsilon$ and adjustment coefficient $w_1$ as $2.5$ and $1.0$, respectively. For our Pani VAT~(+hidden) method, we opt $K_1=10$ and overall perturbation size $\epsilon=2.1$. On the considered two layers, we choose $K_2$ as $10$ and $50$, patch size as $2$ and $1$ and the adjustment coefficient $w$ as $1$ and $4$, respectively.

\paragraph{Our Results.} Table~\ref{table_vat} presents the state-of-the-art performance achieved by  Pani VAT (+hidden)  compared with other baselines on CIFAR-10. We focus on the baseline methods, especially along the direction of variants of VAT and refer to the results from TNAR~(with generative models) method~\citep{yu2019tangent}, the previous state-of-the-art variant of VAT that additionally leverages the data manifold by generative models to decompose the directions of virtual adversarial smoothness. It is worthy to remark that the performance of relevant GAN-based approaches, such as Localized GAN~(LGAN)~\citep{qi2018global} as well as TNAR~(with generative models) in Table~\ref{table_vat}, heavily rely on the established data manifold by the generative models.  It is well-known that one might come across practical difficulties while implementing and deploying these generative models. By contrast, \textbf{without the requirement of generative models}, our approach can eliminate this disturbance and can still outperform these baselines. In addition, our Pani VAT~(+hidden) achieves slight improvement compared with Pani VAT~(input), which serves as an ablation study, and thus verifies the superiority of manifold regularization mentioned in our Pani framework part. Overall, the desirable flexibility along with desirable stability~(lower standard deviation shown in Table~\ref{table_vat}) of Pani VAT further demonstrates the effectiveness of our Pani strategy.

\begin{figure}[t!]
	\centering\includegraphics[width=0.65\textwidth,trim=50 5 50 5,clip]{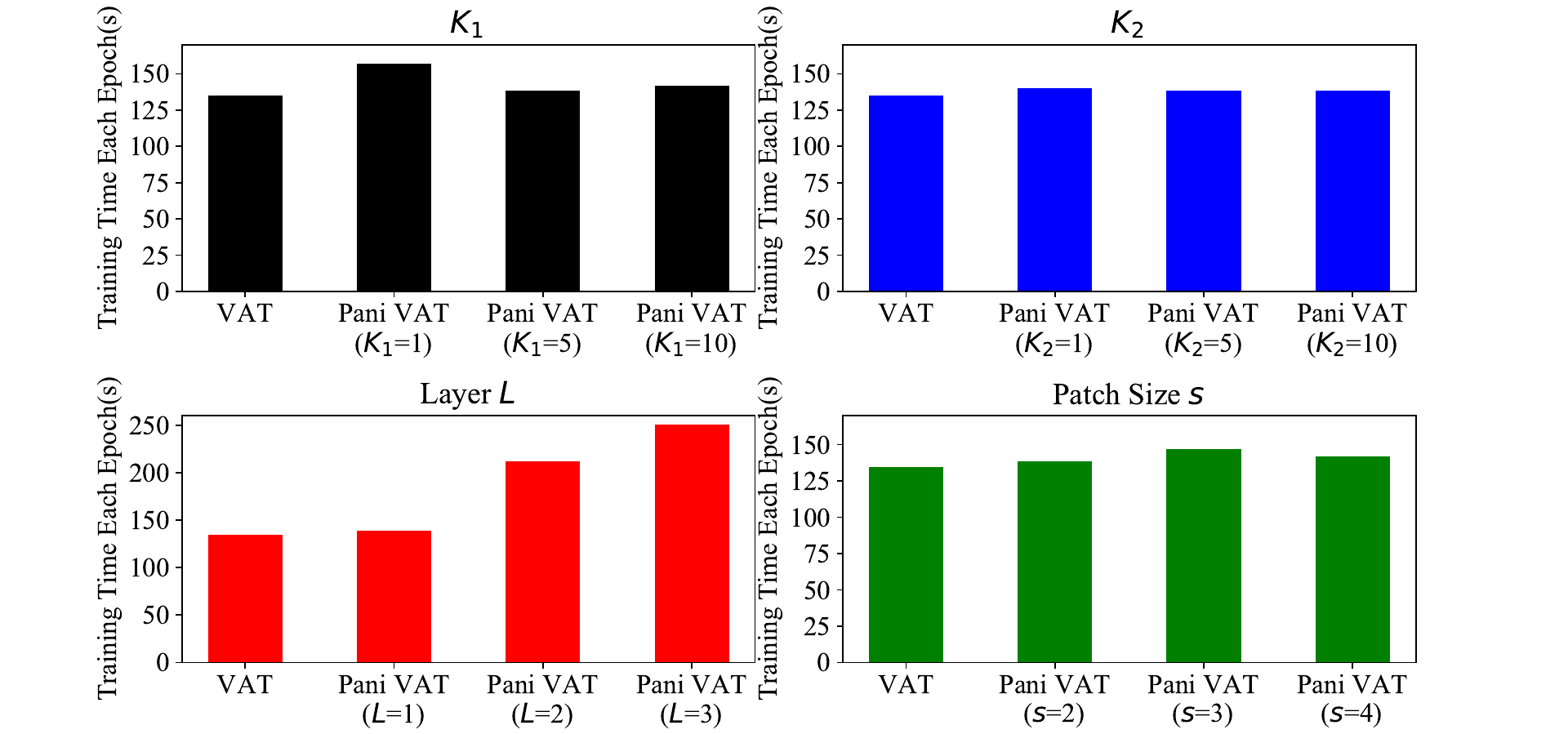}
	\caption{Average training time each epoch with respect to parameters $K_1$, $K_2$, number of layers $L$ and patch size.}
	\label{figure_vat}
\end{figure}

\paragraph{Analysis of Computational Cost.} Another noticeable advantage of our approach is the negligible increase in computation cost compared with Vanilla VAT. In particular, one crucial operation in our approach is the construction of patch set $\mathcal{P}$ and it can be accomplished efficiently by $as.strided$ function in Python or through the specific convolution operation in Pytorch or TensorFlow. The index of $K$ nearest neighbor graph can be efficiently attained through $\textit{topk}$ operation. We conduct further sensitivity analysis on the computational cost of our method concerning other parameters, i.e., $K_1$~(number of peer images), $K_2$~(number of patch neighbors), $L$~(number of perturbed layers) and patch size $s$. As shown in Figure~\ref{figure_vat}, the variation of all parameters has negligible impact on the training time each epoch compared with Vanilla VAT except the number of perturbed layers. The increasing of computational cost presents an almost linear tendency with the increasing of the number of the perturbed layers as the amount of floating-point calculation is proportional to the number of perturbation elements, i.e., $\eta$,  if we temporarily neglect the difference of time in the back propagation process for different layers. Combining results from Table~\ref{table_vat} and Figure~\ref{figure_vat}, we argue that better performance can be expected if we construct perturbations on more hidden layers despite the increase of computation.

\subsection{Pani MixUp}

\paragraph{Implementation Details.} We strictly follow the codebase of MixUp~\citep{zhang2017mixup}, Co-MixUp~\citep{kim2021co} MixMatch~\citep{berthelot2019mixmatch} for a fair comparison, respectively, on CIFAR-10, CIFAR-100 and TinyImageNet datasets. After the line search of hyper-parameters for the best performance, we choose patch size as 16, parameter $a$ in Beta distribution as 2.0 for the data augmentation setting while we choose the patch size 8, $a=2.5$ on the setting without data augmentation across all neural architectures on CIFAR-10 and CIFAR-100. On the TinyImageNet dataset, we set the image dimension as $64\times64$ and batch size as 100. We also introduce a mask mechanism on $\eta$ to avoid overfitting, which randomly set $\eta_{ipk}=0$ based on a ratio. In practice, we set the mask ratio as 0.6 in the data augmentation setting while setting is as 0.4 in the scenario without data augmentation.

\begin{table*}[t!]
	\centering
	\scalebox{0.55}{
		\begin{tabular}{lccccc}
			\toprule[1pt]
			\textbf{Dataset} & \textbf{Model} & \textbf{Aug} & \textbf{ERM} & \textbf{MixUp~($a=1$)} & \textbf{Pani MixUp}~(input)\\
			\hline
			\multirow{6}*{CIFAR-10} &\multirow{2}*{PreAct ResNet-18}&$\checkmark$&5.43 $\pm$ 0.16 &4.24 $\pm$ 0.16 &\textbf{3.93 $\pm$ 0.12}\\
			~&~&$\times$&12.81 $\pm$ 0.46 &9.88 $\pm$ 0.25 &\textbf{8.12 $\pm$ 0.09}\\
			~&\multirow{2}*{PreActResNet-34}&$\checkmark$&5.15 $\pm$ 0.12&3.72 $\pm$ 0.20&\textbf{3.36 $\pm$ 0.15}\\
			~&~&$\times$&12.67 $\pm$ 0.26&10.60 $\pm$ 0.57&\textbf{8.13 $\pm$ 0.32}\\
			~&\multirow{2}*{WideResNet-28-10}&$\checkmark$&4.59 $\pm$ 0.06&3.21 $\pm$ 0.13&\textbf{3.02 $\pm$ 0.11}\\
			~&~&$\times$&8.78 $\pm$ 0.20&8.08 $\pm$ 0.39&\textbf{5.79 $\pm$ 0.03}\\
			\hline
			\multirow{6}*{CIFAR-100} &\multirow{2}*{PreAct	 ResNet-18}&$\checkmark$&24.96 $\pm$ 0.51&22.15 $\pm$ 0.72&\textbf{20.90 $\pm$ 0.21}\\
			~&~&$\times$&39.64 $\pm$ 0.65&41.96 $\pm$ 0.27&\textbf{32.03 $\pm$ 0.34}\\
			~&\multirow{2}*{PreActResNet-34}&$\checkmark$&24.85 $\pm$ 0.14&21.49 $\pm$ 0.68&\textbf{19.46 $\pm$ 0.29}\\
			~&~&$\times$&39.41 $\pm$ 0.80&41.96 $\pm$ 0.24&\textbf{34.48 $\pm$ 0.86}\\
			~&\multirow{2}*{WideResNet-28-10}&$\checkmark$&21.00 $\pm$ 0.09&18.58 $\pm$ 0.16&\textbf{17.39 $\pm$ 0.16}\\
			~&~&$\times$&31.91 $\pm$ 0.77&35.16 $\pm$ 0.33&\textbf{27.71 $\pm$ 0.63}\\
			\hline
			\multirow{6}*{TinyImageNet} &\multirow{2}*{PreAct ResNet-18}&$\checkmark$&44.90 $\pm$ 0.28 &42.84 $\pm$ 0.35&\textbf{42.20 $\pm$ 0.39}\\
			~&~&$\times$&54.95 $\pm$ 0.63&60.58 $\pm$ 0.83&\textbf{52.25 $\pm$ 0.75}\\
			~&\multirow{2}*{PreActResNet-34}&$\checkmark$&40.66 $\pm$ 1.64&43.18 $\pm$ 0.84&\textbf{40.03 $\pm$ 0.61}\\
			~&~&$\times$&51.03 $\pm$ 0.57&55.91 $\pm$ 1.09 &\textbf{49.56 $\pm$ 0.94}\\
			~&\multirow{2}*{WideResNet-28-10}&$\checkmark$& 42.30 $\pm$ 0.51&40.64 $\pm$ 0.77&\textbf{38.97 $\pm$ 0.81}\\
			~&~&$\times$&48.47 $\pm$ 0.24&51.19 $\pm$ 1.19&\textbf{46.26 $\pm$ 0.70}\\
			\bottomrule[1pt]
		\end{tabular}
	}
	\caption{Test error in comparison with ERM, MixUp and Pani MixUp~(input). All  results are averaged under 5 runs. Our implementation is based on MixUp~\citep{zhang2017mixup}. }
	\label{table_mixup}
\end{table*}

\paragraph{Result 1: Comparison with MixUp.} Based on the codebase of MixUp, we compare ERM~(Empirical Risk Minimization), MixUp training and our approach for different neural architectures. For a fair comparison with input MixUp, we conduct our approach only on the input layer. Table~\ref{table_mixup} presents the consistent superiority of Pani MixUp over ERM~(normal training) as well as Vanilla MixUp over different deep neural network architectures. It is worth noting that the superiority of our approach in the setting without data augmentation can be more easily observed than that with data augmentation. Another interesting phenomenon is that MixUp suffers from one kind of collapse in performance as the accuracy of MixUp is even inferior to the ERM on CIFAR-100 and TinyImageNet on the setting without data augmentation. By contrast, our approach exhibits consistent advantages ERM and MixUp across various settings and network architectures.

\begin{table*}[b!]
	\centering
	\scalebox{0.6}{
		\begin{tabular}{lccccccc}
			\hline \text { Dataset (Model) } & \text { Vanilla } & \text { Input } & \text { Manifold } & \text { CutMix } & \text { Puzzle Mix } & \text { Co-MixUp } & \text { Pani MixUp } \\
			\hline \text { CIFAR-100 (PreActResNet18) } & 23.59 & 22.43 & 21.64 & 21.29 & 20.62 & \textbf{19.87} & 20.90\\
			\text { CIFAR-100 (WRN16-8) } & 21.70 & 20.08 & 20.55 & 20.14 & 19.24 & 19.15 & \textbf{19.10}\\
			\text { CIFAR-100 (ResNeXt29-4-24) } & 21.79 & 21.70 & 22.28 & 21.86 & 21.12 & \textbf{19.78} & 21.13  \\
			\text { Tiny-ImageNet (PreActResNet18) } & 43.40 & 43.48 & 40.76 & 43.11 & 36.52 & \textbf{35.85} &  40.19\\
			\hline
		\end{tabular} 
	}
	\caption{Test Error compared with more variants of MixUp, including CutMix and Co-MixUp on CIFAR-100 and Tiny-ImageNet. Results are averaged over 3 seeds after 1,200 epochs. Our implementation is based on Co-MixUp~\citep{kim2021co}. }
	\label{table_mixup_sota}
\end{table*}

\paragraph{Result 2: Comparison with Variants of MixUp.} We empirically demonstrate that Pani MixUp can achieve comparable accuracy compared with variants of MixUp methods, e.g., Co-MixUp. We implement our Pani MixUp method based on the code base of \citep{kim2021co} with the step size 0.05 and additionally compare the performance on WRN16-8 and ResNeXt29-4-24 architectures on CIFAR-100 as opposed to Table~\ref{table_mixup}. In Table~\ref{table_mixup_sota}, all results are evaluated with the data augmentation. It suggests that Pani MixUp outperforms its similar baseline CutMix across all considered settings, which is also based on patches. Our Pani MixUp is also on par with the state-of-the-art Co-MixUp on CIFAR-100, although Co-MixUp performs better than other baselines, especially on Tiny-ImageNet. However, we next show our method is more efficient in computation than Co-MixUp.

\begin{figure*}[t!]
	\centering
	\centering\includegraphics[width=0.7\textwidth,trim=10 80 10 60,clip]{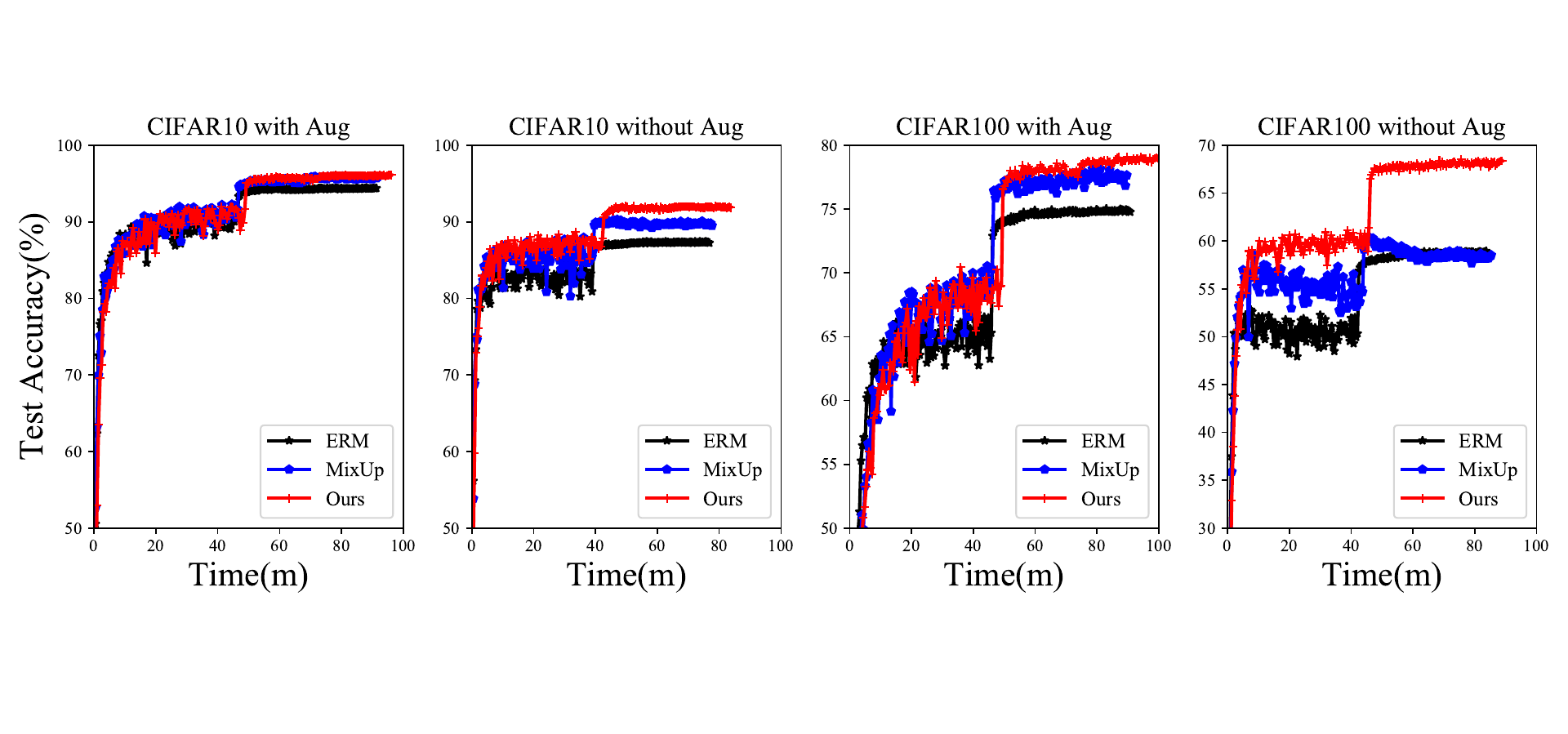}
	\caption{Learning curves with respect to the training time over ERM, MixUp and our approach, where $m$ indicates minutes,  ``with Aug'' and ``without Aug'' denote the settings with and without data augmentation, respectively. }
	\label{figure_mixup}
\end{figure*}

\begin{wrapfigure}[12]{r}{0.4\textwidth}
	\includegraphics[width=0.4\textwidth,trim=0 0 0 0,clip]{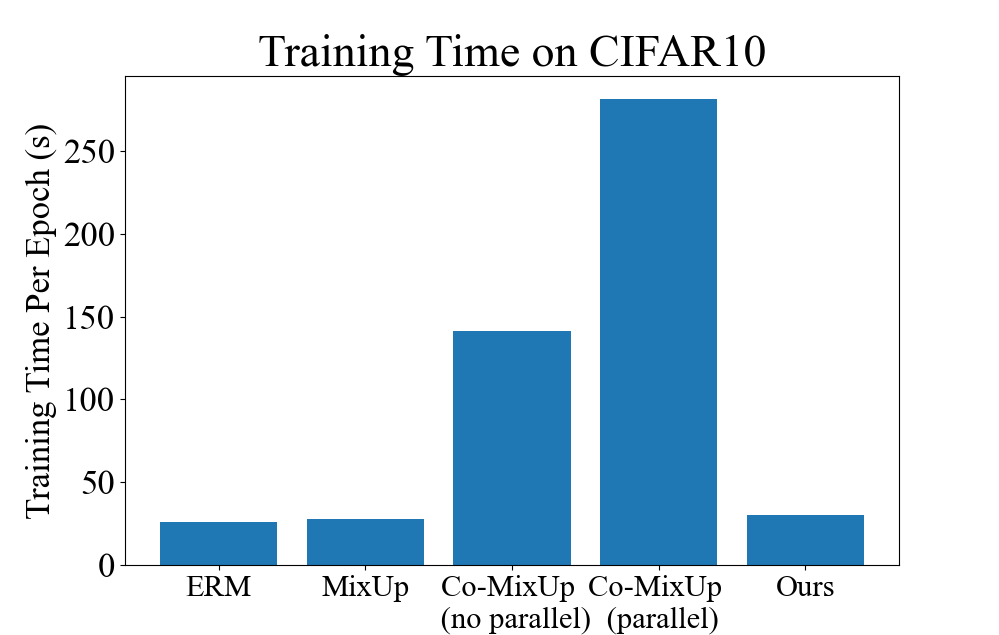}
	\caption{Comparison of Training Time per epoch on CIFAR-10. }
	\label{figure_mixup_computation}
\end{wrapfigure}

\paragraph{Pani MixUp is Computationally Efficient.} We first validate Pani MixUp is as computationally efficient as vanilla MixUp when compared with the normal training~(ERM). Figure~\ref{figure_mixup} presents all learning curves concerning the training time, where we choose ResNet-18 as the basic test model. We can easily observe that with the negligible increase in computation overhead, Pani MixUp achieves better performance than MixUp. In addition, we also compare the computation cost of Pani MixUp with variants of MixUp, especially the state-of-the-art Co-MixUp. In Figure~\ref{figure_mixup_computation}, we reveal that Co-MixUp is more expensive in computation even in the parallel version, although it can achieve favorable performance as suggested in Table~\ref{table_mixup_sota}. By contrast, Pani MixUp achieves comparable results with Co-MixUp on a large number of settings in Table~\ref{table_mixup_sota}, which enjoys significant advantage of the efficiency in computation.

\begin{table}[b!]
	\centering
	\scalebox{0.6}{
		\begin{tabular}{lr}
			\toprule[1pt]
			Methods &  \makecell{CIFAR-10~(4,000 labels)}\\
			\hline
			PiModel~\citep{laine2016temporal}			& $17.41 \pm0.37$ \\ %
			PseudoLabel~\citep{lee2013pseudo}		& $16.21\pm0.11$\\ %
			MixUp~\citep{zhang2017mixup}			& $13.15\pm0.20$\\ %
			VAT~\citep{miyato2017virtual}				& $11.05\pm0.31$\\ %
			MeanTeacher~\citep{tarvainen2017mean}		& $10.36\pm0.25$\\ %
			MixMatch~\citep{berthelot2019mixmatch}		& $6.24\pm0.06$\\ %
			\hline
			Pani MixMatch& $\mathbf{6.08\pm 0.074}$\\
			\bottomrule[1pt]
		\end{tabular}
	}
	\vspace{1.5ex}
	\caption{Performance of our Pani MixMatch in semi-supervised setting on CIFAR-10 with 4000 labels. The reported result is evaluated via the median of last 20 epoch while training average under 4 runs.}
	\label{table_mixmatch}
\end{table}

\paragraph{Further Extension to MixMatch.} We further incorporate our approach into MixMatch~\citep{berthelot2019mixmatch} that naturally extends MixUp to the semi-supervised setting. The resulting approach, which we call Pani MixMatch, elegantly replaces the MixUp part in the MixMatch with our Pani MixUp, thus imposing Pani MixUp by additionally incorporating patch neighborhood correlation knowledge. We apply the same hyper-parameters involved in Pani as in Pani MixUp and other ones are the same as MixMatch. Results in Table~\ref{table_mixmatch} reveals that  Pani MixMatch can further improve the performance of MixMatch in the standard semi-supervised setting, thus verifying the effectiveness and flexibility of our Patch-level Neighborhood Interpolation method.

\section{Discussion and Conclusion}

Since the proposed Pani framework is general and flexible, more regularizations and applications could be considered in the future, such as more regularization methods and applications in natural language processing tasks. We leave this exploration as future works. In addition, an analysis of the theoretical properties of Pani is also valuable in the future.

The recent tendency of the regularization design attaches more importance to consistency and flexibility in various kinds of settings. Along this way, we propose a general regularization motivated by additional leverage of neighboring information existing in the sub-group of samples, e.g., within one batch, which can elegantly extend previous prestigious regularization approaches in a wider range of scenarios. Our proposed Patch-level Neighborhood Interpolation~(Pani) method is flexible and efficient, which can be further incorporated in VAT and MixUp as well as its variants. Our work paves the way toward better understanding and leveraging the knowledge of relationships between samples to design better regularization and improve generalization over a wide range of settings. 

\section*{Acknowledgements}


Z. Lin was supported by National Key R$\&$D Program of China (2022ZD0160302), the major key project of PCL, China (No. PCL2021A12), the NSF China (No. 62276004), and Qualcomm.

\bibliography{Pani}

\end{document}